\documentclass{ceurart}

\usepackage{microtype}

\usepackage{listings}

\begin{document}

\copyrightyear{2022}
\copyrightclause{Copyright for this paper by its authors.
  Use permitted under Creative Commons License Attribution 4.0
  International (CC BY 4.0).}

\conference{CHR 2022: Computational Humanities Research Conference, December 12--14,
  2022, Antwerp, Belgium}

\title{Boosting Word Frequencies in Authorship Attribution}

\author[1]{Maciej Eder}[%
orcid=0000-0002-1429-5036,
email=maciej.eder@ijp.pan.pl,
url=https://maciejeder.org/,
]
\address[1]{Institute of Polish Language, Polish Academy of Sciences,
  al. Mickiewicza 31, 31–120 Kraków, Poland}

\begin{abstract}
In this paper, I introduce a simple method of computing relative word frequencies for authorship attribution and similar stylometric tasks. Rather than computing relative frequencies as the number of occurrences of a given word divided by the total number of tokens in a text, I argue that a more efficient normalization factor is the total number of \textit{relevant} tokens only. The notion of relevant words includes synonyms and, usually, a few dozen other words in some ways semantically similar to a word in question. To determine such a semantic background, one of word embedding models can be used. The proposed method outperforms classical most-frequent-word approaches substantially, usually by a few percentage points depending on the input settings.
\end{abstract}


\maketitle

\section{Introduction}

In a vast majority of text classification studies aimed at distinguishing unique authorial signal -- these include authorship attribution investigations, authorship profiling, verification, and similar tasks -- relative frequencies of the most frequent words (MFWs) are routinely used as the language features to betray the authorial “fingerprint”. A vector of such relative word frequencies is then passed to one of the multidimensional machine-learning classification techniques, ranging from simple distance-based lazy learners, such as Delta \cite{burrowsDeltaMeasureStylistic2002,evertUnderstandingExplainingDelta2017}, to sophisticated deep learning neural network setups \cite{gomez-adornoDocumentEmbeddingsLearned2018}.

Recent advances in machine learning methodology -- unheard-of and unprecedented -- massively reshaped the field of text classification. Three main methodological directions are actively researched: firstly, new classifiers emerge on the horizon to clearly outperform classical solutions; secondly, feature engineering and dimensionality reduction techniques are introduced to overcome the curse of high dimensionality, and thirdly, alternative style-markers that can betray authorial idiosyncrasies are being introduced. The present paper explores none of the above directions, though. Instead, I argue that a reasonable amount of overlooked stylistic information resides in a time-proven, standard bag-of-word representation of textual data, which is routinely used in dozens of stylometric studies. 

Certainly, there exist alternative features that prove to be efficient style-markers in authorship attribution setups. Most notably, letter \textit{n}-grams have been suggested as a strong authorial indicator \cite{pengLanguageIndependentAuthorship2003}. Also, grammatical features, such as POS-tag \textit{n}-grams, turned out to retain information about authorial uniqueness \cite{hirstBigramsSyntacticLabels2007}. Other intriguing ideas include observing the immediate lexical context around proper nouns \cite{lucicSyntacticCharacterizationAuthorship2013}. Even if such alternative textual features exhibit a great deal of potential to enhance text classification \cite{ederStylemarkersAuthorshipAttribution2011}, the standard approach relying of word frequencies continues to be predominant in the field \cite{grieveQuantitativeAuthorshipAttribution2007,stamatatosSurveyModernAuthorship2009}. In this paper, word frequencies will be used as well, yet the step of normalizing them into \textit{relative} frequencies will be somewhat enhanced. Specifically, all the other words used to normalize the frequencies will be evaluated and then reduced, so that a given word in question is normalized by its actual semantic background. However, the general idea of enhancing the frequencies can be extended, I believe, to other style-markers, including extra-lexical ones.

\section{Word frequencies}

The notion of relative word frequencies is fairly simple. We count all the tokens belonging to particular types (e.g. all the attestations of the word “the”, followed by the attestations of “in”, “for”, “of” etc.), and for each word, we divide the number of types by the total number of words in a document. Consequently, each word frequency is equal to its percentage within the document (e.g. “the” = 0.0382), and all the frequencies sum up to 1. The reason of converting occurrences to relative frequencies is obvious: by doing so, one is able to reliably compare texts that differ in length. The notion of relative word frequencies is so natural and intuitive that one might very easily overlook its methodological implications, as if it was nothing else than a simple normalization procedure. 

For the sake of this paper, however, it is important to realize that relative frequencies are relative to \textit{all the other words} in a document in question. Convenient as they are, these values are at the same time very small and – importantly – they are affected by hundreds of other word frequencies. Consequently, the final values might not be sufficiently precise to capture minute differences between word occurrences, because the normalization factor evens them out to some extent. Now, what if we disregard thousands of other words in a text, and instead compute the frequencies in relation to a small number of words that are \textit{relevant}\/? An obvious example is the mutual relation between the words “on” and “upon” in one document \cite{mostellerInferenceDisputedAuthorship1964}; essentially, more attestations of “upon” come at the cost of the occurrences of the word “on” – and vice versa. While the classical relative frequency of the word “on” in Emily Bronte’s \textit{Wuthering Heights} is 0.00687, the proportion of “on” relative exclusively to “upon” is 0.9762. It is assumed in this paper that the latter frequency can betray the authorial signal to a greater extent than the classical approach, because the myriads of other words are not blurring the final value.

The idea of looking into semantics is not entirely new, since thesaurus-based approaches have been already proposed in the context of authorship attribution  \cite{loveAttributingAuthorshipIntroduction2002,koppelFeatureInstabilityCriterion2006}. It has been suggested that a list of words organized into near-synonymous sets (“synsets”) and/or into larger hierarchies can be used to extract the authorial signal \cite{juolaThesaurusbasedSemanticSimilarity2018}, it has been also demonstrated that pairs of synonyms might contain valuable authorial information \cite{borskiCopernicusHisLatin2021}. However, the above approaches are focused on identifying meaningful words \textit{beyond} the usual MFWs, whereas the present study is aimed to show that there is still some room to enhance the very MFWs.

\section{Method}

Given the above “on” and “upon” example, it would be tempting to identify one synonym for each of the words, and to compute the relative proportions in each of the synonym pairs, as suggested in the already cited study \cite{borskiCopernicusHisLatin2021}. Linguistically speaking, however, such an approach would hardly be feasible. Firstly, only a fraction of words have their proper synonyms. Secondly, some semantic fields are rather rich and cannot be reduced to a mere pair of synonyms. Thirdly, in the case of the most frequent words (articles, particles, prepositions) identifying their synonyms doesn’t make much sense, yet still, \textit{relevant counterparts} for these frequent words obviously exist. On theoretical grounds, however, it is difficult to speculate whether the number of relevant counterparts should be restricted to a single word – as in the example of “on” defined by its relation to “upon” – or include, say, a dozen related words. E.g., to determine the relative frequency of the word “make”, one would probably measure its proportion against the sum of occurrences of “do”, “prepare”, “create”, “turn”, “craft”, “invent” etc. The effective size of the semantic background is, however, very difficult to conceptualize -- not only the actual number of related words, but even the order of magnitude are unknown. Take the above example: should the word “make” be calculated against its 10 similar words, or would the semantic background of 100 words be better?

\begin{figure*}
  \centering
  \includegraphics[width=\linewidth]{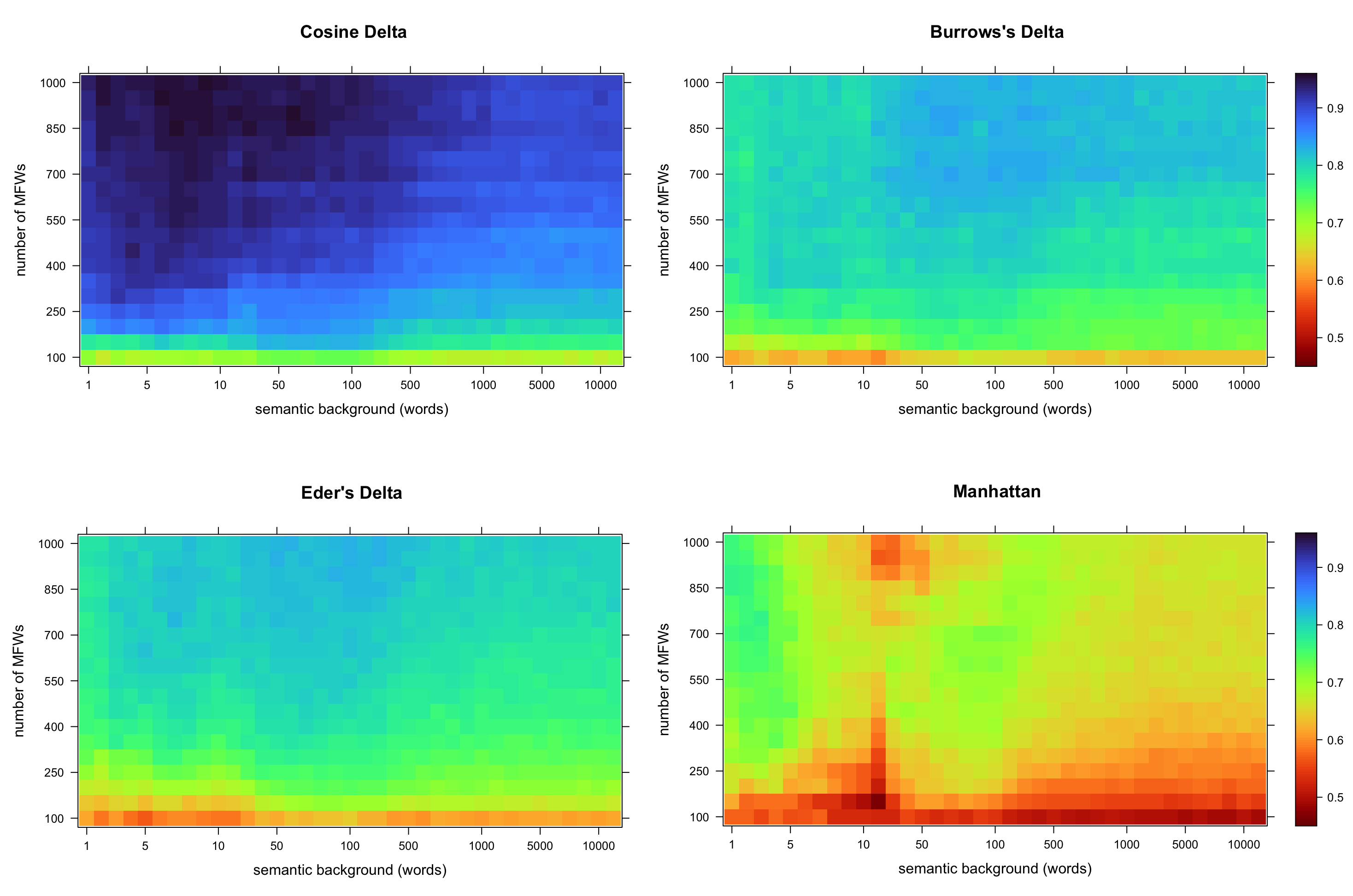}
    \caption{The performance (F1 scores) for a benchmark corpus of 99 English novels, and the Delta classifier. Distance measures involve Cosine Delta (top left), Classic Delta (top right), Eder's Delta (bottom left), and Manhattan (bottom right). The results depend on the MFW vector (\textit{y} axis) and the size of the semantic space expressed in the number of most similar words in a vector model (\textit{x} axis).}
\end{figure*}

\begin{figure*}
  \centering
  \includegraphics[width=\linewidth]{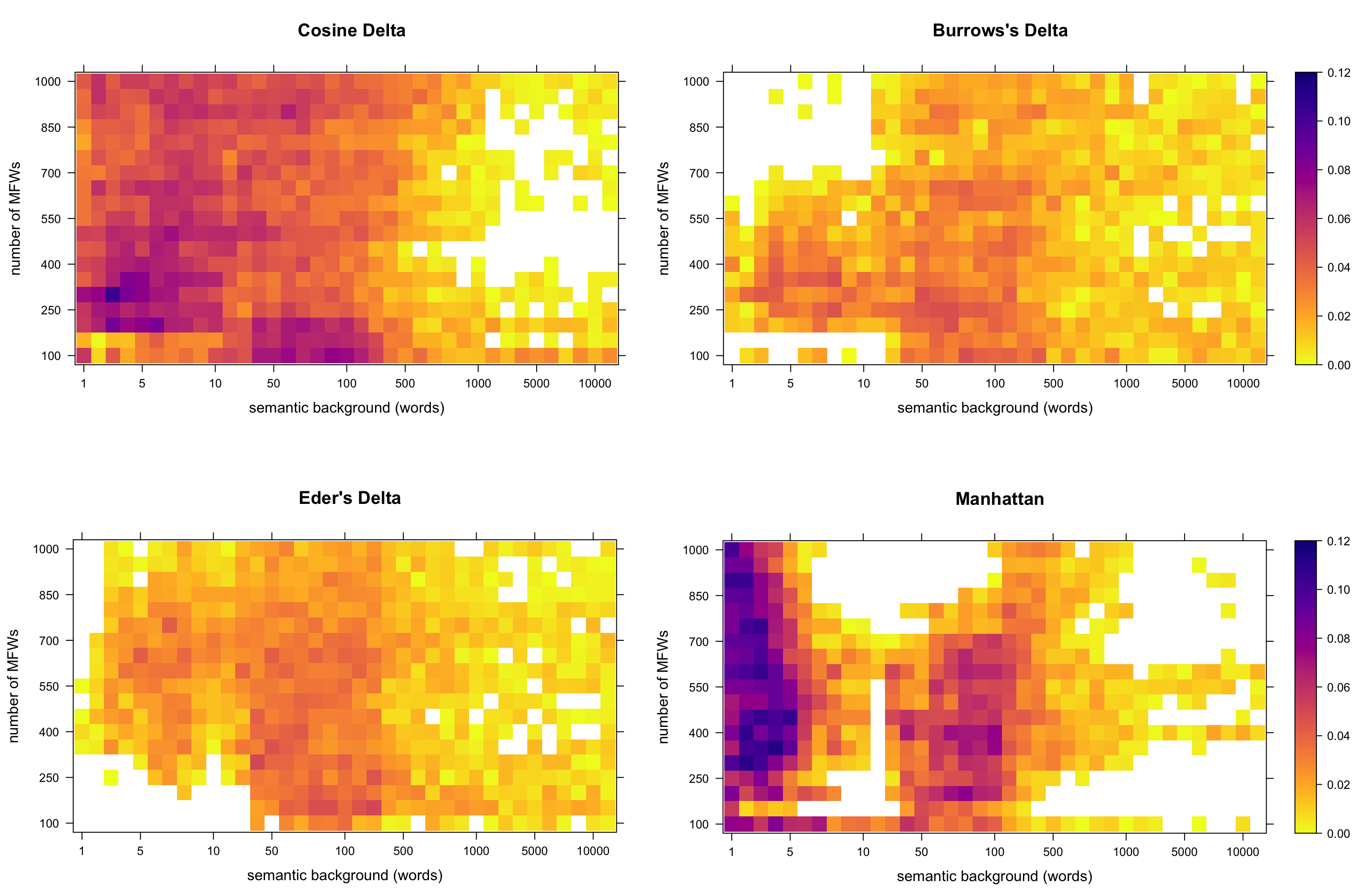}
    \caption{The gain of performance (baseline F1 scores subtracted from the obtained F1 scores) for a benchmark corpus of 99 English novels.}
\end{figure*}

\begin{figure*}
  \centering
  \includegraphics[width=\linewidth]{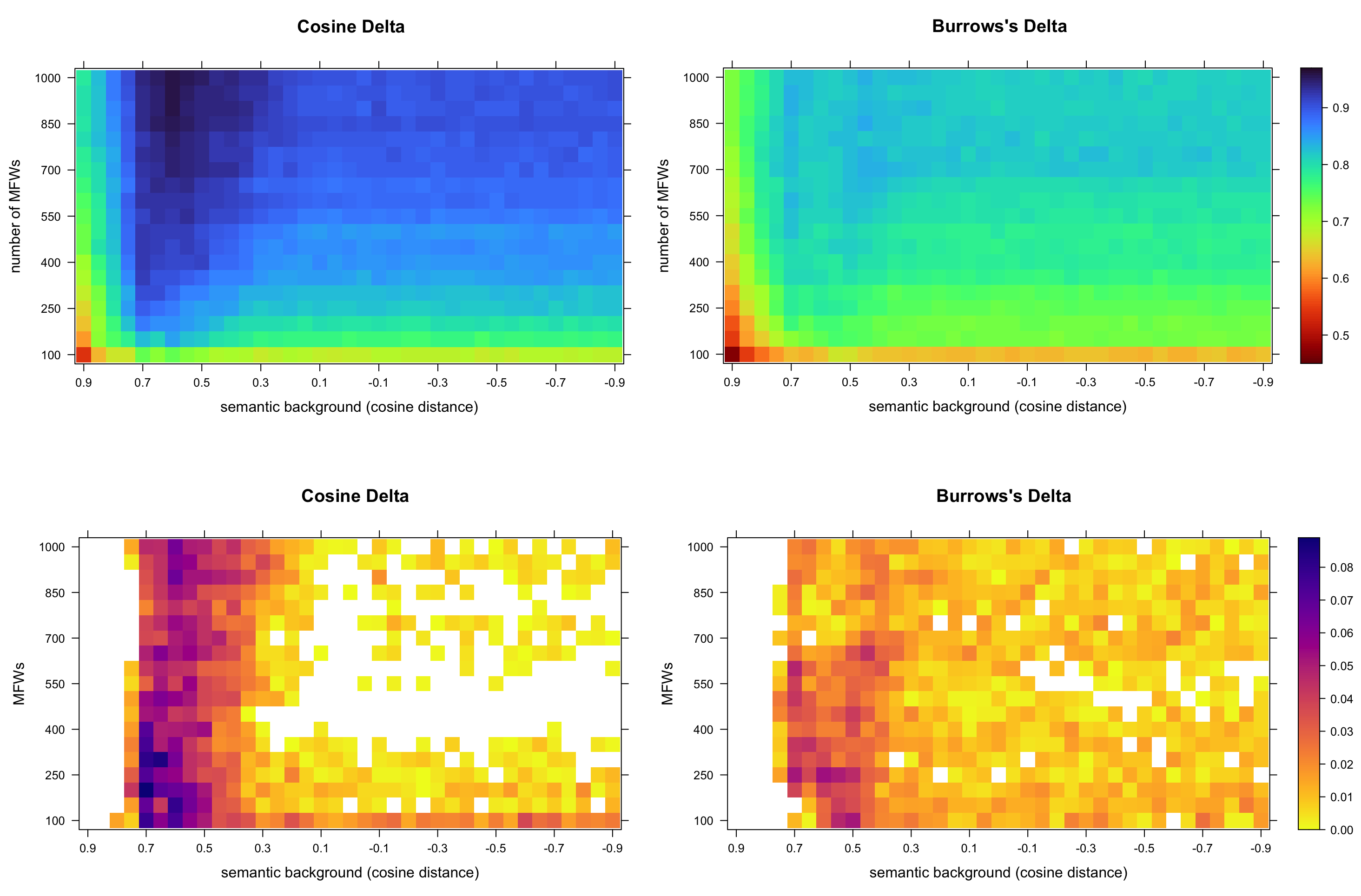}
    \caption{The absolute performance (top) and the performance gain (bottom) for the corpus of 99 English novels. The distance measures are Cosine Delta (left) and Burrows’s Delta (right). The semantic space is defined as the words within a given cosine distance from respective source words. The results depend on the MFW vector (\textit{y} axis) and the size of the semantic space (\textit{x} axis).}
\end{figure*}

Another nontrivial question is related to the very method of extracting synonyms and other semantically related words from a corpus. While thesaurus-based search might prove feasible for single words, it will certainly become more demanding when dozens of seed words are concerned. There exist, however, at least two strategies to approach the issue computationally. One strategy involves \texttt{wordnet}, a manually compiled database of thousands of words with their semantic and syntactic relations \cite{millerWordNetLexicalDatabase1995}, while the other relies on distributional semantics methods. In particular, the algorithm \texttt{word2vec} should be mentioned in this context  
\cite{mikolovDistributedRepresentationsWords2013}, which provides a vector representation of words that allows for identifying their semantic similarities. Even if these inferred similarities do not comply with any formal grammar (rather, the relations are known to be fuzzy at times), they usually look convincing to a human observer. In the present study, a word vector model \texttt{GloVe} \cite{penningtonGloVeGlobalVectors2014} was used to betray word similarities: it was trained on the benchmark corpus of 99 English novels (as described below), with 100 target dimensions. A semantic background for a given seed word was defined as \textit{n} neighboring vectors. Consequently, the resulting semantic background contained the most similar vectors for a given seed word. E.g., the neighbors for the word “person” were: “woman”, “gentleman”, “man”, “one”, “sort”, “whom”, “thing”, “young”, etc., whereas the neighbors for the word “the” were as follows: “of”, “this”, “in”, “there”, “on”, “one”, “which”, “its”, “was”, “a”, “and”, etc. For each target word, a relative frequency was calculated as the number of occurrences divided by the sum of occurrences of its \textit{n} semantic neighbors (\textit{n} being the size of semantic space to be tested).

In order to corroborate the above intuitions, a controlled authorship attribution experiment was designed. A benchmark corpus of 99 English novels was used: it consists of 33 authorial classes and 3 novels per author, and is freely available on GitHub repository: \url{https://github.com/computationalstylistics/100_english_novels}. A corpus of (naturally long) novels might be considered inferior for authorship benchmarks, the high number of the authorial classes, however, makes the task difficult enough to sufficiently stress-test the classifier. To make the task even harder, the amount of training data were restricted to 1 text per author, whereas the remaining 2 texts per author were used as the validation set (the proportion of 33 \textit{vs}. 66 texts were kept in each iteration). 

Since the size of the semantic background is unknown, a grid-search framework was designed to systematically assess tighter (1 relevant counterpart) and broader semantic spaces (up to 10,000 words, inevitably going far beyond synonyms). The tests were performed using the package \texttt{stylo} for R \cite{ederStylometryPackageComputational2016}. Different classifiers, MFW vectors and, most importantly, different sizes of the semantic space were tested systematically, in a supervised setup with stratified cross-validation. On theoretical grounds, the size of the semantic space \textit{n} = 80,000 (roughly the total number word types in the benchmark corpus) would be equivalent to classical relative frequencies, whereas the space of the size \textit{n} = 1 means that the frequencies are relative to exactly one other word (e.g. the frequency of the word “the” would be the number of occurrences of “the” divided by the total number of “the” and “of”).

Independently, an alternative set of tests were performed using a regular relative frequencies. The outcomes of these tests served as a baseline. In each test reported in this paper, the F1 scores are uses as a compact and reliable measure of performance.

\section{Results}

The obtained results (Table 1, and Fig. 1) clearly suggest that the new method outperforms the classical relative frequencies solution \textit{substantially}, no matter which distance measure is used. In agreement with several previous studies, longer MFW vectors worked better than, say, 100 MFWs. Also according to expectation, Cosine Delta proves to be the undisputed winner among the classifiers. Counter-intuitive, however, was the behavior of different classifiers with the enhanced word frequencies. As evidenced in Fig. 1 top left panel, Cosine Delta works best with frequencies computed against 5–50 semantically similar words, whereas Burrows’s Delta (top right) exhibits its sweet spot for 50–100 neighboring words, and so does Eder’s Delta (bottom left). When the semantic background is further increased, the behavior of particular classifiers becomes uniform across the board: it slowly but surely decreases to ultimately reach the baseline level. 

Since the introduction of Burrows’s Delta practitioners are aware that scaling (\textit{z}-scoring) the features is the very factor responsible for the performance boost observed in Delta and its derivatives. Even if Manhattan distance does not scale the features (hence its unpopularity in text classification), the improved word frequencies behave differently than standard approaches, which in turn might favor simple distances such as Manhattan. And indeed, the scores obtained for the Manhattan Distance are radically better than the respective baseline (Fig. 1, bottom right), yet still, Manhattan still cannot compete with \textit{z}-scored distances.    

According to the above results, a recipe for a successful authorship attribution setup seems to be as follows: take roughly 800–900 MFWs, and compute their frequencies using, for each word, the occurrences of their 5–10 semantic neighbors; then use the Cosine Delta classifier.


Since in authorship attribution the results are proven to be unevenly distributed across different MFW vectors, let alone different classifiers, Fig. 2 presents the same outcomes as previously, yet this time defined as the improvement (in percentage points) over the baseline F1 scores. While the overall best performance is obtained for \textit{ca}. 850 MFWs computed against 5–10 words, the biggest gain over the baseline (more than 10 percentage points!) is provided by the following scenario: 300 MFW frequencies computed against a tight semantic background of 3 neighboring words. Other reasonable improvements are generally associated with short MFW vectors and the semantic background of 5–100 words. In the case of Burrows’s Delta, which worked with 900 MFWs computed against 60 neighboring words (Fig. 2, top right), the improvement over the baseline is biggest for short vectors of MFWs. Interestingly, for Burrows’s Delta the new method proves to be \textit{worse} than the baseline for long MFW vectors and tight semantic spaces of 1–10 neighboring words. The picture for Eder’s Delta (bottom left) is similar to that for Burrows’s method, even if its hot spot is slightly moved towards longer MFWs vectors. Surprisingly enough, the results for Manhattan Distance turned out to be substantially different from the other methods, and much less predictable. A large and pronounced hot spot of radically improved performance forms for tight semantic spaces, across different MFWs vectors. On the right hand side, the mountain of performance is followed by a deep valley of no improvement at all, and then, counter-intuitively, another hill emerges, indicating the boost of performance for the semantic spaces of 50–100 words. This behavior is difficult to explain.

\begin{table*}
    \caption{The best performance (F1 scores) obtained in each tested scenario.}
  \label{tab:freq}
  \begin{tabular}{ccl}
    \toprule
         & Relative frequencies & Enhanced frequencies\\
    \midrule
    Cosine Delta    & 0.908 & \textbf{0.959} \\
    Burrows’s Delta & 0.823 & 0.838 \\
    Eder’s Delta    & 0.812 & 0.830 \\
    Manhattan       & 0.679 & 0.771 \\
  \bottomrule
\end{tabular}
\end{table*}

The proposed way of identifying an arbitrarily chosen number of semantic neighbors, might suffer from an uneven distribution of semantic neighbors in a given model (GloVe, word2vec, fastText, etc.). E.g., 50 neighboring lexemes might point to a semantically coherent area around a function word, or indicate but vague associations around a very specific technical term. To account for this factor, a second experiment has been conducted, in which I have defined a semantic background to be all the words located at a specific cosine distance from a given reference word. Consequently, rather than extracting \textit{n} neighboring words, now I was extracting all the words within the radius of 0.9 cosine similarity in the first iteration, then 0.85, 0.8 etc., all the way to –0.9. The results for Cosine Delta and Burrows’s Delta are shown in Fig. 3. As can be seen, a clear hot spot forms in the area of 0.7–0.5 cosine similarity, despite the number of MFWs or the classifier, and after the distance of 0.3 the performance hits the baseline level. The results confirm the general picture obtained in the previous experiment (Fig. 1), yet the sweet spot area seems to be more difficult to generalize.

\section{Discussion}

The results presented in the previous section call for further exploration and above all, for a concise discussion. A few general remarks can be formulated here:

\begin{enumerate}
  \item No matter which classification method was used, the performance improvement turned out to be large, clearly suggesting that bare word occurrences retain much more authorial signal than the time-proven relative frequencies are able to betray. It can be safely hypothesized that the method introduced in this paper barely opened a new perspective, rather than offered an ultimate solution to the problem.
  \item In order to identify \textit{the words that matter}, a word embedding model was used – and this, again, was far from an optimal solution. As a rough proxy, it nevertheless was able to improve the word frequencies in the range of 5–50 neighboring words. On theoretical grounds, a further improvement should be possible with a more precise method of identifying relevant semantic background. 
  \item While the new method improves the performance across all the MFW strata, short MFW vectors seem to benefit more. Interesting from a theoretical point of view, this phenomenon has also a practical implication. Namely, since several studies suggest that larger numbers of MFWs should be preferred as they generally exhibit better performance, it is also believed that they are more likely to be affected by genre, topic, and content of the analyzed texts. With this in mind, some practitioners choose to conduct authorship attribution on shorter MFW vectors. The method introduced in this paper can greatly improve the performance in such setups.
\end{enumerate}

An observation that requires further investigation, is the discrepancy between classifiers in how they react to the same semantic background. Contrary to intuition, for Burrows’s Delta the improvement of performance was not simply correlated with the size of the semantic background. Tight neighborhood – less than 20 synonyms and/or other related words – did not outperform standard relative frequencies, whereas broader contextual information of \textit{ca}. 50–100 related words showed a significant improvement over the baseline. In the case of Cosine Delta, tight semantic background of \textit{ca}. 5–10 proved optimal, whereas broader spaces of 50–100 neighboring words were only marginally worse, still outperforming the baseline to a significant degree.


\section{Conclusion}

The paper presented a simple method to improve the performance in different stylometric setups. The method is conceptually straightforward and does not require any NLP tooling. The only external piece of information that is required is a list of semantically related words for each of the most frequent words in the corpus. A controlled experiment showed a significant improvement of classification accuracy in a supervised multi-class authorship attribution setup.

\begin{acknowledgments}
  This research is part of the project \textit{Large-Scale Text Analysis and Methodological Foundations of Computational Stylistics} (2017/26/E/HS2/01019), supported by Poland’s National Science Centre. The code and the datasets to replicate the experiments presented in this study are posted on GitHub repository: \url{https://github.com/computationalstylistics/word_frequencies}.

\end{acknowledgments}

\bibliography{bibliography.bib}


\appendix

\section{Function to compute enhanced word frequencies}

The following code defines a function to compute the word frequencies as discussed in this paper. The code is written in generic R and does not require any external R library to run. The function takes three arguments: (i) \texttt{word\_frequencies} is a document-term matrix, or a table with raw frequencies (occurrences) or words in a given dataset; unlike typical stylometric applications, where one usually takes a subset of \textit{n} most frequent words, here all the information about infrequent words is equally important; (ii) \texttt{word\_vector\_similarities} is a table containing, for each word, the nearest neighbors in a semantic space, e.g. the row for the word “person” these are the following words: “woman”, “gentleman”, “man”, “one”, “sort”, “whom”, “thing”, “young”, etc.; it is sufficient to compute the neighbors for 1000 most frequent words or so, and the semantic depth can be reduced to, say, 100 semantically related words in each case (for the sake of the present study, a set of 1000 most frequent words with their 10,000 semantic neighbors were used); (iii) \texttt{no\_of\_similar\_words} a number (integer) of how many semantic neighbors one wants to take into consideration.

\begin{lstlisting}
compute_subset_frequencies = function(dtm_matrix, 
                                    word_vector_similarities, 
                                    no_of_similar_words) {

  semantic_space = word_vector_similarities[ , 1:no_of_similar_words, 
                     drop = FALSE]
  no_of_words = dim(semantic_space)[1]
  final_frequency_matrix = matrix(nrow = dim(dtm_matrix)[1], 
                     ncol = no_of_words)

  for(i in 1:no_of_words) {

      # check if the required word(s) appears in the corpus
      words_sanitize = semantic_space[i,] %in% colnames(dtm_matrix)
      words_to_compute = semantic_space[i, words_sanitize]
      # if the corpus doesn't contain any of the words required 
      # by the model, then grab the the most frequent word
      # for reference (it should not happen often, though)
      if(length(words_to_compute) == 0) {
        words_to_compute = colnames(dtm_matrix)[1]
      }
      # add the occurences of the current word being computed;
      # e.g. for the word "of", add "of" to the equation
      words_to_compute = c(colnames(dtm_matrix)[i], words_to_compute)
      # getting the occurrences of the relevant words from
      # the input matrix of word occurrences:
    f = dtm_matrix[, words_to_compute]
    # finally, computing new relative frequencies
    final_frequency_matrix[,i] = f[,1] / rowSums(f)

  }

  # sanitizing again, by replacing NaN values with Os
  final_frequency_matrix[is.nan(final_frequency_matrix)] = 0
  # tweaking the names of the rows and columns
  rownames(final_frequency_matrix) = rownames(dtm_matrix)
  colnames(final_frequency_matrix) = rownames(semantic_space)
  class(final_frequency_matrix) = "stylo.data"

return(final_frequency_matrix)
}
\end{lstlisting}

\end{document}